\documentclass{article}

\usepackage{microtype}
\usepackage{graphicx}
\usepackage{subcaption}
\usepackage{booktabs} 

\usepackage{hyperref}
\usepackage[most]{tcolorbox}
\usepackage{enumitem}
\usepackage{amsfonts}       
\usepackage{nicefrac}       
\usepackage{subcaption}
\usepackage{listings}
\usepackage{minted}

\newcommand{\code}[1]{\texttt{\hyphenchar\font=45\relax #1}}

\usepackage[preprint]{icml2026}

\usepackage{amsmath}
\usepackage{amssymb}
\usepackage{mathtools}
\usepackage{amsthm}
\usepackage{multicol}
\usepackage{multirow}
\usepackage{xcolor}
\usepackage{colortbl}
\usepackage[dvipsnames]{xcolor}

\definecolor{citecolor}{HTML}{0071BC}
\definecolor{linkcolor}{HTML}{ED1C24}
\definecolor{grey}{HTML}{999999}
\definecolor{green}{HTML}{ABD1BC}
\definecolor{lightblue}{HTML}{B0C4DE}
\definecolor{purple}{HTML}{E3BBED}
\definecolor{orange}{HTML}{ffdab9}

\usepackage[capitalize,noabbrev]{cleveref}

\theoremstyle{plain}

\theoremstyle{definition}

\theoremstyle{remark}

\usepackage[textsize=tiny]{todonotes}

\icmltitlerunning{SAIL-RL: Guiding MLLMs in When and How to Think via Dual-Reward RL Tuning}

\begin{document}

\twocolumn[
  \icmltitle{SAIL-RL: Guiding MLLMs in When and How to Think via \\ Dual-Reward RL Tuning}



  \icmlsetsymbol{equal}{*}

  \begin{icmlauthorlist}
    \icmlauthor{Fangxun Shu}{equal,bytedance}
    \icmlauthor{Yngjie Ye}{equal,bytedance}
    \icmlauthor{Yue Liao}{equal,nus}
    \icmlauthor{Zijian Kang}{bytedance}
    \icmlauthor{Weijie Yin}{bytedance} 
    \icmlauthor{Jiacong Wang}{bytedance}
    \icmlauthor{Gengyuan Liu}{bytedance}
    \icmlauthor{Xiao Liang}{bytedance}
    \icmlauthor{Shuicheng Yan}{nus}
    \icmlauthor{Chao Feng}{bytedance}
  \end{icmlauthorlist}

  \icmlaffiliation{bytedance}{Douyin SAIL Team}
  \icmlaffiliation{nus}{National University of Singapore}

  \icmlcorrespondingauthor{Shuicheng Yan}{yansc@nus.edu.sg}
  \icmlcorrespondingauthor{Xiao Liang}{liangxiao.ilx@bytedance.com}
  \icmlcorrespondingauthor{Chao Feng}{chaofeng.zz@bytedance.com}

  \icmlkeywords{Machine Learning, ICML}

  \vskip 0.3in
  ]



\printAffiliationsAndNotice{\icmlEqualContribution}

\begin{abstract}
We introduce SAIL-RL, a reinforcement learning (RL) post-training framework that enhances the reasoning capabilities of multimodal large language models (MLLMs) by teaching them when and how to think. Existing approaches are limited by outcome-only supervision, which rewards correct answers without ensuring sound reasoning, and by uniform thinking strategies, which often lead to overthinking on simple tasks and underthinking on complex ones. SAIL-RL addresses these challenges with a dual reward system: the Thinking Reward, which evaluates reasoning quality through factual grounding, logical coherence, and answer consistency, and the Judging Reward, which adaptively determines whether deep reasoning or direct answering is appropriate. Experiments on the state-of-the-art SAIL-VL2 show that SAIL-RL improves reasoning and multimodal understanding benchmarks at both 4B and 8B scales, achieving competitive performance against commercial closed-source models such as GPT-4o, and substantially reduces hallucinations, establishing it as a principled framework for building more reliable and adaptive MLLMs. The code will be available at \url{https://github.com/BytedanceDouyinContent/SAIL-RL}

\begin{figure}[t]
    \centering
    \includegraphics[width=0.48\textwidth]{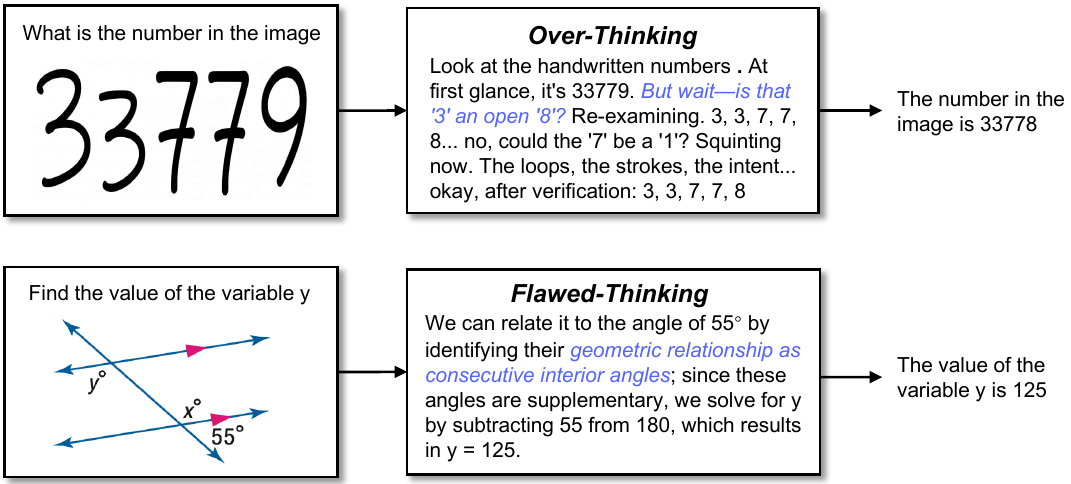}
   
    \caption{Limitations of current MLLMs in reasoning. \textbf{Top:} Overthinking where the model applies a needlessly complex reasoning process to a simple problem, resulting in an incorrect answer. \textbf{Bottom:} Lucky success where the model reaches the correct answer through a flawed reasoning process. 
}
    \label{fig:outcome_only_supervision}
\end{figure}

\end{abstract}
\section{Introduction}
Multimodal Large Language Models (MLLMs)~\citep{gemini,gpt-4o,qwen25vl,llava,internvl} are rapidly advancing from basic visual recognition toward complex reasoning and holistic understanding. This evolution is fundamentally driven by the refinement of training paradigms that define how models internalize intelligence from large-scale multimodal data. While Supervised Fine-Tuning (SFT) established the foundation for multimodal instruction following, the field has increasingly shifted its focus toward hybrid post-training frameworks~\citep{kimivl,wethink} that integrate SFT with Reinforcement Learning (RL). Notably, RL is undergoing a pivotal paradigm shift: moving beyond mere alignment with human preferences~\citep{openairlhf,dpo,kto}, recent methodologies~\citep{openvlthinker, wethink,vlaathinker,keyevl} commonly follow the paradigm of \textit{``thinking before speaking.''} Guided by a special token \texttt{\textbackslash think}, the model first generates a structured reasoning trace before producing the final answer. Leveraging long reasoning chains as an internal knowledge source allows the model to extract salient cues that improve answer accuracy and strengthen overall capability. Nevertheless, despite these advances, current methods still face two fundamental challenges:

\begin{figure*}[h]
    \begin{center}
    
        \includegraphics[width=0.99\textwidth]{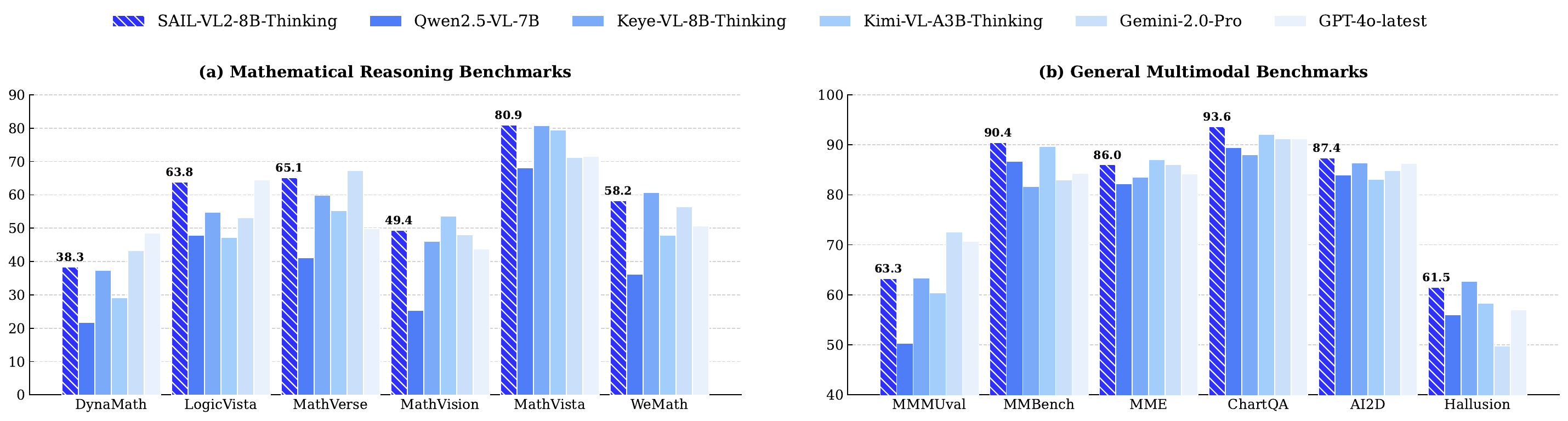} 
        
        \caption{\textbf{Performance comparison between SAIL-VL2-Thinking (SAIL-VL2 post-trained with our SAIL-RL) and other LVMs.}
SAIL-VL2-Thinking achieves superior advantages on both general understanding and mathematical reasoning benchmarks at the 8B scale and delivers competitive performance against large-scale closed-source models.}
        \label{fig:performance}
    \end{center}
\end{figure*}

\underline{Answers without sound reasoning:} Conventional methods rely on \textit{outcome-only supervision}, where rewards are determined by the correctness of the final answer, while the quality of reasoning is ignored. This paradigm introduces two critical issues: first, as the intuition \textit{“think well to answer right”} suggests, incoherent or redundant reasoning traces hinder the model from extracting useful cues, leading to inaccurate answers and exacerbating hallucinations. As shown in Figure~\ref{fig:outcome_only_supervision} Bottom, conventional MLLMs~\citep{keyevl} can produce correct answers despite factual errors in reasoning, highlighting how outcome-only rewards compromise robustness and trustworthiness. Second, during optimization, models may occasionally reach correct answers through flawed or fabricated reasoning paths. Such spurious alignments are nevertheless reinforced as positive outcomes, fostering a form of “false correctness” that undermines both robustness and reliability.

\underline{Overthinking the easy, underthinking the hard:} Most approaches apply the same reasoning process to all tasks, regardless of complexity. This uniformity often leads to overthinking on simple problems, introducing unnecessary cost and noisy reasoning chains. As illustrated in Figure~\ref{fig:outcome_only_supervision} Top, models frequently generate redundant reasoning for trivial queries (\emph{e.g.}, object color recognition), highlighting the inefficiency of static strategies. Conversely, on complex problems, the same rigidity causes underthinking, producing shallow reasoning and inaccurate answers. The lack of adaptive control prevents models from allocating cognitive resources efficiently, unlike humans who naturally adjust their effort based on task difficulty.

To address these challenges, we propose \textbf{SAIL-RL}, a novel post-training framework for MLLMs. While following the standard two-stage paradigm of CoT-augmented SFT and RL-tuning, SAIL-RL introduces a dual reward system that supervises both \textit{reasoning quality} and \textit{reasoning efficiency}. The \textbf{Thinking Reward} moves beyond outcome-only supervision by directly assessing the reasoning process. It evaluates logical coherence to maintain step-by-step validity, factual grounding to mitigate hallucinations, and trace-to-answer consistency to ensure that the final answer is faithfully derived from the reasoning process. The \textbf{Judging Reward} enhances adaptivity by enabling models to decide when deep reasoning is necessary. The model learns to adopt a direct-answer mode for simple tasks and a full-reasoning mode for complex ones, improving efficiency while aligning cognitive resource allocation more closely with human behavior. Together, these two reward systems allow SAIL-RL to strengthen both the reliability and efficiency of MLLMs in reasoning and comprehensive tasks.

We conduct extensive experiments to evaluate the effectiveness of SAIL-RL. Building on the state-of-the-art MLLM SAIL-VL2, we develop SAIL-VL2-Thinking through our RL-based post-training strategy. As shown in Fig.~\ref{fig:performance}, with the dual reward system, SAIL-VL2-Thinking delivers consistent gains over the baseline and conventional RL-based approaches, achieving state-of-the-art results on multiple reasoning benchmarks at 8B scales. It also reaches leading performance on OpenCompass, maintains competitive accuracy on general multimodal understanding tasks, and substantially reduces hallucinations, highlighting the robustness and reliability introduced by SAIL-RL. Together, these contributions establish SAIL-RL as a principled post-training framework that strengthens both the quality and adaptivity of reasoning in MLLMs.

\begin{figure*}[t]
\begin{center}
    \includegraphics[width=0.98\textwidth]{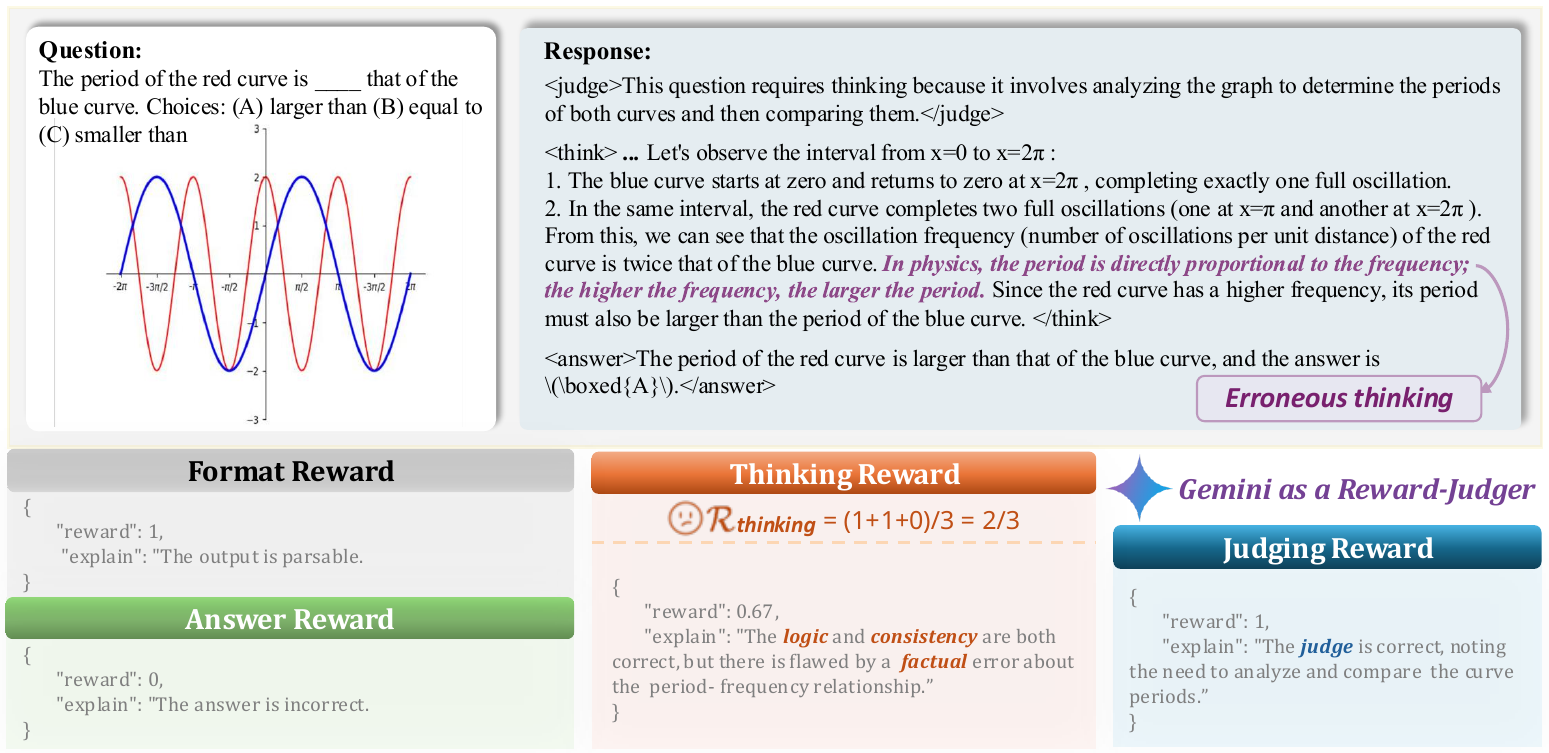} 
\vspace{-2mm}
\caption{An overview of the SAIL-RL's reward system. The system evaluates a model's response across four dimensions: Format, Answer, Thinking, and Judging. The nuanced semantic rewards for Thinking and Judging are provided by Gemini acting as a reward-judger.}
\vspace{-5mm}
\label{fig:framework}
\end{center}
\end{figure*}
\section{Related Work}

\textbf{Cognitive Paradigms in MLLMs.} Recent advancements in MLLMs can be categorized into two cognitive paradigms: System 1 and System 2. System 1 models, exemplified by state-of-the-art generalist models~\citep{gpt-4o, qwen25vl}, prioritize rapid, intuitive visual perception. These models typically utilize direct instruction tuning to map visual features to textual outputs, excelling at perception-intensive tasks but often struggling with complex logical derivations. Conversely, System 2 models aim to emulate slow, deliberative reasoning. Following the breakthrough of DeepSeek-R1~\citep{guo2025deepseekr1} in text reasoning, recent works have extended this paradigm to the multimodal domain. Models such as Gemini-2.5-Pro~\citep{gemini} and Kimi-VL-Thinking~\citep{kimivl} leverage reinforcement learning to internalize extensive Chain-of-Thought (CoT) processes. By generating explicit thinking tokens before the final answer, these models achieve significant gains in complex visual math and logic tasks. However, the high inference cost of System 2 reasoning necessitates dynamic architectures. Recent works~\citep{zhang2025othink, zhou2025autovla} have begun exploring routing mechanisms to switch between these modes, yet optimizing when and how to switch remains an open challenge.

\textbf{Multimodal RL.} RL has emerged as a powerful paradigm for enhancing long-chain reasoning in LLMs~\citep{gpto1,guo2025deepseekr1}. Leveraging this foundation, recent research has extended RL to MLLMs~\citep{vlaathinker,keyevl,openvlthinker,wang2025vl}, achieving significant improvements in visual reasoning. However, directly applying these RL methods to multimodal tasks introduces two critical challenges: efficiency bottlenecks caused by overthinking simple multimodal questions, and effectiveness bottlenecks caused by incorrect multimodal reasoning chains. Consequently, current optimization strategies largely focus on addressing these symptoms in isolation. To mitigate efficiency issues, works such as R-4B employ a bi-mode policy optimization to determine whether to activate the reasoning process. To improve reasoning quality, VisualPRM~\citep{wang2503visualprm} and URSA~\citep{luounlocking} utilize visual process reward models for step-by-step reasoning verification. In contrast to these approaches that typically optimize dimensions in isolation, we aim to design a dual-reward mechanism, which simultaneously learns when and how to reason, unifying efficiency and effectiveness within a single framework.

\section{SAIL-RL}

We propose SAIL-RL to improve both the effectiveness and efficiency of RL for MLLMs. SAIL-RL introduces a dual-reward mechanism that guides models on \textit{how} and \textit{when to think}. Specifically, our design features a Thinking Reward to guarantee reasoning quality and a Judging Reward to enable adaptive reasoning. Crucially, these are unified via a Cascading Reward System, a multiplicative formulation that ensures both rewards work in strict synergy.

\subsection{Thinking Reward: How to Think}
As the saying goes, \emph{``sound reasoning leads to correct answers.''} To improve response quality, a model is required to learn \emph{how to think} by constructing clear and coherent reasoning paths. Beyond outcome-only supervision in conventional RL-tuning, we introduce a \textbf{Thinking Reward} that comprehensively evaluates reasoning quality with LLM-based judge models. This reward is integrated into RL tuning to guide models toward producing higher-quality reasoning across multiple dimensions.

\textbf{Logical Coherence Reward.} We first introduce the Logical Coherence Reward, which evaluates whether a model can \textit{think clearly}. This dimension measures the internal logical integrity of the reasoning process, ensuring arguments are both well-structured and meticulously executed. To this end, the judge model applies two sequential checks: (i) \textit{Structural Soundness}, assessing whether the problem is properly decomposed and formulated (e.g., into valid equations or logical steps); and (ii) \textit{Deductive Soundness}, verifying that subsequent steps are free of contradictions, calculation errors, or logical fallacies. Failure in either check yields a score of $d_1 = 0$, while success in both yields $d_1 = 1$.

\textbf{Factual Grounding Reward.} We then introduce the Factual Grounding Reward to evaluate whether the model is \textit{thinking truthfully} rather than hallucinating. This reward penalizes unsupported statements by requiring each step in the reasoning process to be factually grounded. To this end, the judge model performs a hierarchical fact-check across three sources: (i) \textit{Visual Grounding}, verifying claims against the provided image; (ii) \textit{Textual Grounding}, checking consistency with the input query; and (iii) \textit{World Knowledge}, consulted only when verification is not possible from the first two sources. Any contradiction at any stage yields a score of $d_2 = 0$; otherwise, $d_2 = 1$.

\noindent\textbf{Answer Consistency Reward.}
We further introduce the Answer Consistency Reward to evaluate whether the model \textit{thinks faithfully}. This dimension ensures that the final answer is a direct and logical conclusion derived strictly from the preceding reasoning. The judge model verifies that the reasoning trace fully justifies the answer, explicitly checking for disconnects, reliance on unstated information, or unsupported leaps in logic. Any failure results in a score of $d_3 = 0$; otherwise, the score is $d_3 = 1$.

Finally, the overall thinking reward $\mathcal{R}_{\text{think}}$ is computed as the average of these three dimensions: $\mathcal{R}_{\text{think}} = \frac{1}{3} \sum_{i=1}^{3} d_i$.

\subsection{Judging Reward: When to Think} 
We further guide MLLMs on \textit{when to think}: applying detailed reasoning only for complex problems while giving direct responses to simple ones. The goal is to balance efficiency and effectiveness by adapting the reasoning to task difficulty. To this end, we introduce a \textbf{Judging Reward}, which incentivizes the model to determine whether reasoning is necessary before generating a response.

Specifically, the model is required to output a thinking decision before answering. This decision is evaluated against ground-truth complexity labels. The reward $d_{\text{judge}}$ is binary: it is set to $1$ if the model's decision aligns with the ground truth (i.e., choosing thinking mode for complex tasks or no-thinking mode for simple tasks); otherwise, it is $0$. This simple mechanism penalizes both \textit{under-reasoning} on complex tasks and \textit{over-thinking} on simple tasks. By optimizing $\mathcal{R}_{\text{judge}} = d_{\text{judge}}$, SAIL-RL enables adaptive reasoning, ensuring that rigorous thinking is triggered only when necessary.

\subsection{Cascading Reward System}
\label{sec:cascade_reward}
We formulate the core reasoning signal as a multiplicative interaction between its constituent components. As shown in Eq.~\ref{eq:total_reward}, the joint success of the judging decision ($\mathcal{R}_{\text{judge}}$), the thinking quality ($\mathcal{R}_{\text{think}}$), and the final answer accuracy ($\mathcal{R}_{\text{answer}}$) is integrated via a cascading product:

\vspace{-2mm}
\begin{equation}
\small
\mathcal{R}_{\text{total}} = \alpha \cdot (\mathcal{R}_{\text{judge}} \cdot \mathcal{R}_{\text{think}} \cdot \mathcal{R}_{\text{answer}}) + (1 - \alpha) \cdot \mathcal{R}_{\text{format}},
\label{eq:total_reward}
\end{equation}

\noindent where $\alpha = 0.9$. is a balancing coefficient that prioritizes logical rigor while ensuring adherence to response format. This structure functions as a \textbf{logical AND gate}, ensuring that rewards are granted only when the judgment, reasoning, and answer are all correct. This design imposes a zero-tolerance penalty to prevent reward hacking. For example, if a model attempts a \textit{lucky guess} on a complex problem by skipping reasoning, the term $\mathcal{R}_{\text{judge}}$ becomes 0, nullifying the entire reward even if the answer happens to be correct. Similarly, if it selects thinking but generates irrelevant or empty reasoning, $\mathcal{R}_{\text{think}}$ becomes 0. Note that we also incorporate standard rewards, where $\mathcal{R}_{\text{answer}}$ evaluates response correctness and $\mathcal{R}_{\text{format}}$ acts as a regularizer for structural compliance. This mechanism eliminates the possibility of obtaining rewards through shortcuts, forcing the model to strictly align its reasoning behaviors with the complexity of the task. The prompts of thinking reward and judging reward are provided in Appendix~\ref{appendix:reward_prompt}.

\subsection{Post-training Strategy}

We employ a two-stage post-training strategy to instill and refine the reasoning capabilities of SAIL-RL. We first utilize Supervised Fine-Tuning (SFT) to establish a structured reasoning format, followed by Reinforcement Learning (RL) to optimize the reasoning quality and efficiency using our proposed reward system.

\noindent\textbf{Stage1: LongCoT SFT.} The first stage builds the model's foundational ability to sequentially judge a problem's complexity, generate a step-by-step reasoning process, and derive a final answer.  To support our ``Judge-Think-Answer'' paradigm, we construct a LongCoT dataset where each sample is structured as a sequence of judgment, reasoning, and answer. Specifically, we utilize a strong teacher model to generate a judgment (\verb|<judge>|) on whether the problem requires complex reasoning, followed by a detailed thinking process (\verb|<think>|) leading to the ground-truth answer enclosed in a \verb|\boxed{}| tag. This explicit structure prevents the model from bypassing reasoning on complex tasks. We fine-tune the base model using a standard next-token prediction loss over the full sequence. The objective function is defined as:
\begin{equation}
\small
\label{eq:longcot_sft_loss}
\mathcal{L}_{\text{LongCoT-SFT.}} = - \frac{1}{|D_{\text{CoT}}|} \sum_{(I, J, T, A) \in D_{\text{CoT}}} \log P_\theta(J \circ T \circ A | I)
\end{equation}
where $I$ is the input, $J$ is the judgment text, $T$ the reasoning process, $A$ the final answer, and $\circ$ denotes concatenation. This training objective teaches the model to first judge the problem's nature, then produce a corresponding reasoning trace, and finally output the answer in the correct format.


\noindent\textbf{Stage2: RL Tuning with Reward System.} While LongCoT SFT provides a strong generative template, RL further incentivizes the model to optimize \textit{how to think} (quality) and \textit{when to think} (efficiency).
To ensure stable RL training and prevent reward hacking, we curate a dataset on verifiable tasks. We convert multiple-choice questions into free-response formats and apply difficulty-based filtering to remove trivial or unsolvable samples. This ensures the model learns from problems within an optimal difficulty range. We optimize the SFT model using the DAPO~\citep{dapo} algorithm with our proposed dual-reward system ($\mathcal{R}_{\text{total}}$). We remove the KL divergence penalty adjusts higher clip to encourage exploration.

\section{Experiments} \label{sec:experiments}

In this section, we evaluate SAIL-RL across various vision-language reasoning and understanding benchmarks. Due to space constraints, comprehensive implementation details, including dataset construction, specific hyperparameters and hardware configurations, are provided in Appendix~\ref{appendix:implementation}.

\begin{table*}[t]
  \caption{Evaluation results on OpenCompass multimodal reasoning benchmarks. The best results among open-source models are \textbf{bolded} and the second-best results are \underline{underlined}.}
  \vspace{-3mm}
  \label{reasoning_vl}
  \begin{center}
    \begin{small}
      \begin{sc}
        \resizebox{\textwidth}{!}{%
          \setlength{\tabcolsep}{4pt}
          \begin{tabular}{lcccccc c} 
            \toprule
            \textbf{Model} & \textbf{DynaMath} & \textbf{LogicVista} & \textbf{MathVerse} & \textbf{MathVision} & \textbf{MathVista} & \textbf{WeMath} & \textbf{Average} \\
            \midrule
            \rowcolor{gray!15} \multicolumn{8}{c}{\textit{Close-source Models}} \\
            \midrule
            Gemini-2.0-Pro & 43.3 & 53.2 & 67.3 & 48.1 & 71.3 & 56.5 & 56.6 \\
            GPT-4o-latest & 48.5 & 64.4 & 49.9 & 43.8 & 71.6 & 50.6 & 54.8 \\
            \midrule
            \rowcolor{gray!15} \multicolumn{8}{c}{\textit{Open-source Models}} \\
            \midrule
            InternVL3-2B & 14.0 & 33.6 & 20.6 & 20.2 & 57.3 & 13.0 & 26.5 \\
            Qwen2.5-VL-3B & 11.0 & 36.0 & 29.3 & 18.1 & 60.2 & 20.7 & 29.2 \\
            WeThink-7B & 24.4 & 53.0 & 44.7 & 27.2 & 70.9 & 48.0 & 44.7 \\
            InternVL3-8B & 25.7 & 44.5 & 38.5 & 30.0 & 70.5 & 39.5 & 41.5 \\
            Qwen2.5-VL-7B & 21.8 & 47.9 & 41.1 & 25.4 & 68.1 & 36.2 & 40.1 \\
            VL-Rethinker-7B & 17.8 & 42.7 & 46.4 & 28.4 & 73.7 & 36.3 & 40.9 \\
            VLAA-Thinker-7B & 22.4 & 48.5 & 48.2 & 26.4 & 68.0 & 41.5 & 42.5 \\
            Keye-VL-8B-Thinking & \underline{37.3} & \underline{54.8} & \underline{59.8} & 46.0 & \underline{80.7} & \textbf{60.7} & \underline{56.6} \\
            Kimi-VL-A3B-Thinking & 29.1 & 47.2 & 55.2 & \textbf{53.6} & 79.5 & 47.9 & 52.1 \\
            \midrule 
            SAIL-VL2-2B-Instruct & 10.2 & 36.2 & 22.6 & 23.4 & 71.1 & 22.7 & 31.0 \\
            SAIL-VL2-2B-LongCoT & 18.3 & 38.6 & 41.8 & 27.7 & 72.4 & 35.9 & 39.1 \\
            SAIL-VL2-2B-Thinking & 25.7 & 45.4 & 50.5 & 30.5 & 73.6 & 42.1 & 44.6 \\
            SAIL-VL2-8B-Instruct & 17.8 & 45.0 & 32.9 & 27.6 & 76.4 & 35.8 & 39.3 \\
            SAIL-VL2-8B-LongCoT & 29.7 & 58.2 & 53.1 & 39.7 & 77.2 & 54.4 & 52.1 \\
            SAIL-VL2-8B-Thinking & \textbf{38.3} & \textbf{63.8} & \textbf{65.1} & \underline{49.4} & \textbf{80.9} & \underline{58.2} & \textbf{59.3} \\
            \bottomrule
          \end{tabular}%
        }
      \end{sc}
    \end{small}
  \end{center}
  \vskip -0.1in
\end{table*}

\begin{table*}[h!]
  \caption{Evaluation on multimodal understanding benchmarks. HallBench denotes HallusionBench. The best results among open-source models are \textbf{bolded} and the second-best results are \underline{underlined}.}
   \vspace{-3mm}
  \label{general_vl}
  \begin{center}
    \begin{small}
      \begin{sc}
        \resizebox{1.0\textwidth}{!}{%
          \setlength{\tabcolsep}{4pt}
          \begin{tabular}{lcc cccc cc} 
            \toprule
            \multirow{2}{*}{\textbf{Model}} & \multicolumn{3}{c}{\textbf{General VQA}} & \multicolumn{3}{c}{\textbf{OCR \& Chart}} & \textbf{Hallucination} & \multirow{2}{*}{\textbf{Average}} \\
            \cmidrule(lr){2-4} \cmidrule(lr){5-7} \cmidrule(lr){8-8}
            & \textbf{MMMU} & \textbf{MMBench} & \textbf{MME} & \textbf{ChartQA} & \textbf{AI2D} & \textbf{OCRBench} & \textbf{HallBench} & \\ 
            \midrule
            \rowcolor{gray!15} \multicolumn{9}{c}{{\textit{Close-source Models}}} \\ 
            \midrule
            Gemini-2.0-Pro & 72.6 & 83.0 & 86.1 & 91.2 & 84.8 & 86.3 & 49.8 & 77.9 \\ 
            GPT-4o-latest & 70.7 & 84.3 & 84.2 & 91.5 & 86.3 & 82.2 & 57.0 & 79.0 \\ 
            \midrule
            \rowcolor{gray!15} \multicolumn{9}{c}{{\textit{Open-source Models}}} \\ 
            \midrule
            InternVL3-2B & 47.1 & 84.3 & 77.4 & 80.4 & 78.7 & 83.5 & 41.4 & 68.2 \\ 
            Qwen2.5VL-3B & 48.1 & 82.4 & 77.5 & 87.0 & 80.7 & 79.7 & 48.3 & 70.7 \\ 
            WeThink-7B & 50.9 & 87.8 & 82.9 & 90.8 & 84.5 & 88.9 & 55.1 & 75.3 \\ 
            InternVL3-8B & 57.3 & 87.7 & 85.2 & 89.6 & 85.2 & 88.0 & 53.7 & 76.5 \\ 
            Qwen2.5-VL-7B & 50.3 & 86.7 & 82.2 & 89.5 & 84.0 &  86.4 & 56.0 & 74.8 \\ 
            VL-Rethinker-7B & 54.8 & 88.2 & 82.9 & 91.5 & 83.6 & 89.1 & 55.1 & 76.0 \\ 
            VLAA-Thinker-7B & 51.9 & 86.9 & 83.3 & 89.5 & 78.9 & 89.4 & 51.5 & 73.7 \\ 
            Keye-VL-8B-Thinking* & \underline{63.4} & 81.7 & 83.5 & 88.0 & 86.4 & 85.1 & \textbf{62.7} & 77.6 \\ 
            Kimi-VL-A3B-Thinking* & {60.4} & 89.7 & \textbf{87.0} & \underline{92.1} & 83.1 & 82.3 & {58.3} & \underline{78.4} \\ 
            \midrule
            SAIL-VL2-2B-Instruct & 47.7 & 86.8 & 76.6 & 89.1 & 83.0 & 89.5 & 51.7 & 72.5 \\ 
            SAIL-VL2-2B-LongCoT  & 44.6 & 82.5 & 74.7 & 90.2 & 77.4 & 87.4 & 54.0 & 70.6 \\ 
            SAIL-VL2-2B-Thinking & 51.2 & 87.2 & 78.4 & 92.2 & 84.1 & 90.1 & 53.1 & 74.1 \\ 
            SAIL-VL2-8B-Instruct & 55.4 & \underline{90.2} & 84.5 & 90.3 & \textbf{87.7} & \underline{90.5} & 55.1 & 77.2 \\ 
            SAIL-VL2-8B-LongCoT  & 63.0 & 88.7 & 82.6 & 91.3 & 83.6 & 88.6 & 59.4 & 78.1 \\ 
            SAIL-VL2-8B-Thinking & \textbf{66.1} & \textbf{90.4} & \underline{86.0} & \textbf{93.6} & \underline{87.4} & \textbf{91.3} & \underline{61.5} & \textbf{80.8} \\ 
            \bottomrule
          \end{tabular}%
        }
      \end{sc}
    \end{small}
  \end{center}
  \vskip -0.1in
  \vspace{-3mm}
\end{table*}

\subsection{Experimental Setups}

\noindent \textbf{Model Architecture.} Our model is built on SAIL-VL2~\citep{sailvl2}, which integrates AimV2~\citep{aimv2} and Qwen3~\citep{qwen3}. We employ a two-stage training comprising full-parameter SFT followed by RL using DAPO~\citep{dapo} to optimize reasoning capabilities.

\noindent \textbf{Training Datasets.} The LongCoT SFT stage utilizes 400K high-quality samples in a \texttt{judge-think-answer} format to instill meta-cognitive capabilities. Second, the RL Stage employs a 70K mixed dataset, featuring 50K STEM-focused problems with verifiable reward and 20K general QA samples from LLaVA-OneVision~\citep{llava_onevision}.

\noindent \textbf{Evaluation Benchmarks.} We employ VLMEvalKit~\citep{duan2024vlmevalkit} for evaluation, using GPT-4o-Mini as the model judge. We assess two primary categories of abilities: Advanced Reasoning, evaluated on benchmarks focused on mathematical and logical analysis such as DynaMath, MathVerse, MathVista, and WeMath~\citep{dynamath, zhang2024mathverse, lu2023mathvista, qiao2025wemath}; and General Multimodal Understanding, evaluated using comprehensive benchmarks ranging from general VQA and chart comprehension to hallucination detection, including MMMU~\citep{yue2024mmmu} and MMBench~\citep{liu2024mmbench}.

\subsection{Benchmark Performance}


We evaluate \textbf{SAIL-VL2-8B-Thinking} across various benchmarks in both reasoning and general understanding. Beyond quantitative metrics, we provide qualitative case studies in Appendix~\ref{sec:case_study} to offer deeper insights into the model's emergent reasoning behaviors. These examples specifically contrast \textbf{SAIL-RL}'s logical fidelity against common baseline pitfalls—such as ``accidental correctness'' in math and ``overthinking'' in perception tasks—illustrating the practical impact of our dual-reward system, which are omitted from the main text due to space constraints.

\noindent \textbf{Multimodal Reasoning Benchmarks.} As shown in Table~\ref{reasoning_vl}, SAIL-VL2-8B-Thinking achieves a new state-of-the-art among open-source models with an average score of 59.3. This performance leap is primarily driven by our thinking reward, which shifts the focus from simple outcome matching to supervising the underlying quality of the reasoning process. By incentivizing logical coherence and rigorous step-by-step deduction, the model achieves a +20.0 improvement over the SAIL-VL2-8B baseline (39.3). The enhanced logical depth is particularly evident in reasoning-intensive tasks, where our model delivers top-tier results on DynaMath (38.3), LogicVista (63.8), and MathVista (80.9), notably surpassing frontier closed-source systems such as GPT-4o (54.8) and Gemini-2.0-Pro (56.6).

\noindent \textbf{Multimodal Understanding Benchmarks.} As shown in Table~\ref{general_vl}, SAIL-VL2-8B-Thinking achieves an open-source state-of-the-art average score of 80.4. Beyond deep reasoning, the judging reward plays a critical role in general tasks by preventing ``overthinking'' on straightforward perception problems. By identifying simple tasks and bypassing unnecessary reasoning steps, the model significantly reduces the potential for hallucinations---a common pitfall in forced thinking models---as evidenced by its 61.5 on HallusionBench and 93.6 on ChartQA. These results demonstrate that SAIL-RL effectively balances reasoning depth with computational efficiency, enabling robust performance across varied task complexities.

\subsection{Thinking Reward Enhances Reasoning Capability}
To validate the effectiveness of the thinking reward, we conduct an ablation study using SAIL-VL2-8B trained with only the answer reward as our baseline. As shown in Table~\ref{tab:ablation_stem_only}, the addition of the thinking reward yields substantial gains in reasoning-intensive tasks, including improvements of +3.2 on WeMath, +1.6 on DynaMath, +2.4 on LogicVista, and +2.1 on MathVision. These results indicate that thinking reward effectively steers the model from superficial outcome-matching toward a deeper exploration of underlying logical structures, thereby enhancing its robustness in high-dimensional reasoning spaces.

\begin{figure}[t] 
    \centering
    \includegraphics[width=\columnwidth]{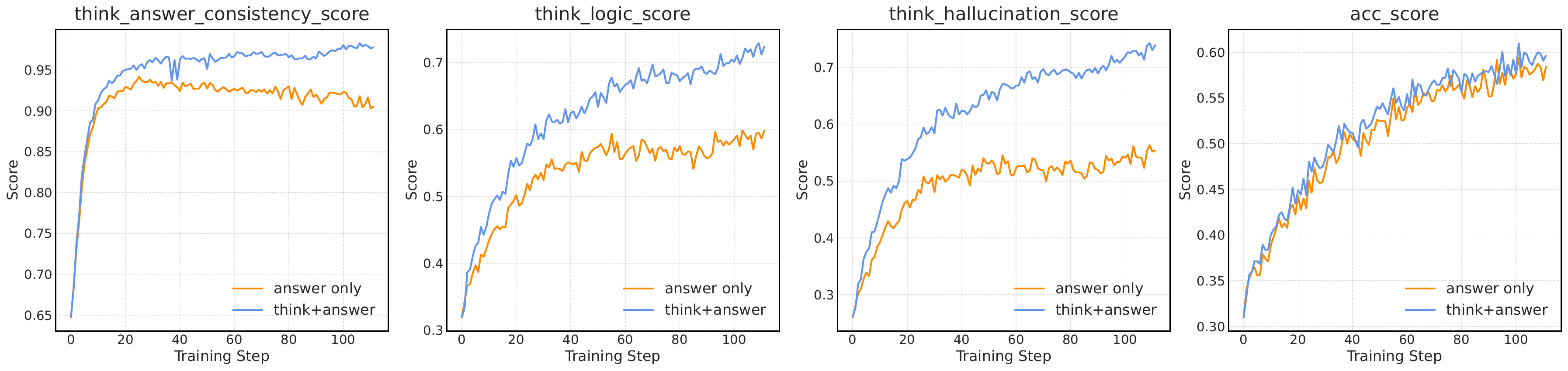}
    \caption{Training dynamics of thinking reward. SAIL-RL (blue) consistently improves all three thinking score over the answer-only baseline (orange), which stagnates or degrades.}
   \label{fig:training_dynamics}
\end{figure}

Analysis of the training dynamics further elucidates the evolution of internal reasoning logic. As illustrated in Figure~\ref{fig:training_dynamics}, both the logic score and the hallucination mitigation score (blue curves) exhibit a steady and synchronized ascent throughout the training duration, while the answer-only baseline (orange curves) plateaus early. Most critically, the consistency score highlights a fundamental divergence: the thinking reward maintains near-perfect alignment between the reasoning chain and the final answer, whereas the baseline—lacking intermediate process constraints—suffers from a distinct "reasoning collapse" in later training stages. This phenomenon, characterized by the progressive decoupling of the thinking process from the final output, demonstrates that outcome-only supervision is insufficient to sustain coherent logical support.

\begin{table}[h!]
  \centering
  \caption{Performance comparison of Thinking Reward on STEM benchmarks. SAIL-RL (answer+thinking reward) consistently outperforms the answer-only baseline.}
  \label{tab:ablation_stem_only}
  
  \begin{small}
    \begin{sc}
      \resizebox{\columnwidth}{!}{%
        \setlength{\tabcolsep}{3pt}
        \begin{tabular}{l cccc}
          \toprule
          \textbf{Reward Type} & \textbf{WeMath} & \textbf{LogicVista} & \textbf{MathVision} & \textbf{DynaMath} \\
          \midrule
          Answer & 55.0 & 61.4 & 47.3 & 36.7 \\
          \textbf{SAIL-RL} & \textbf{58.2} & \textbf{63.8} & \textbf{49.4} & \textbf{38.3} \\
          \midrule
          \rowcolor[gray]{.95} \textit{Improvement} & \textit{+3.2} & \textit{+2.4} & \textit{+2.1} & \textit{+1.6} \\
          \bottomrule
        \end{tabular}%
      }
    \end{sc}
  \end{small}
  \vskip -0.1in
 
\end{table}

\begin{figure}[t]
\begin{center}
    \includegraphics[width=\columnwidth]
    {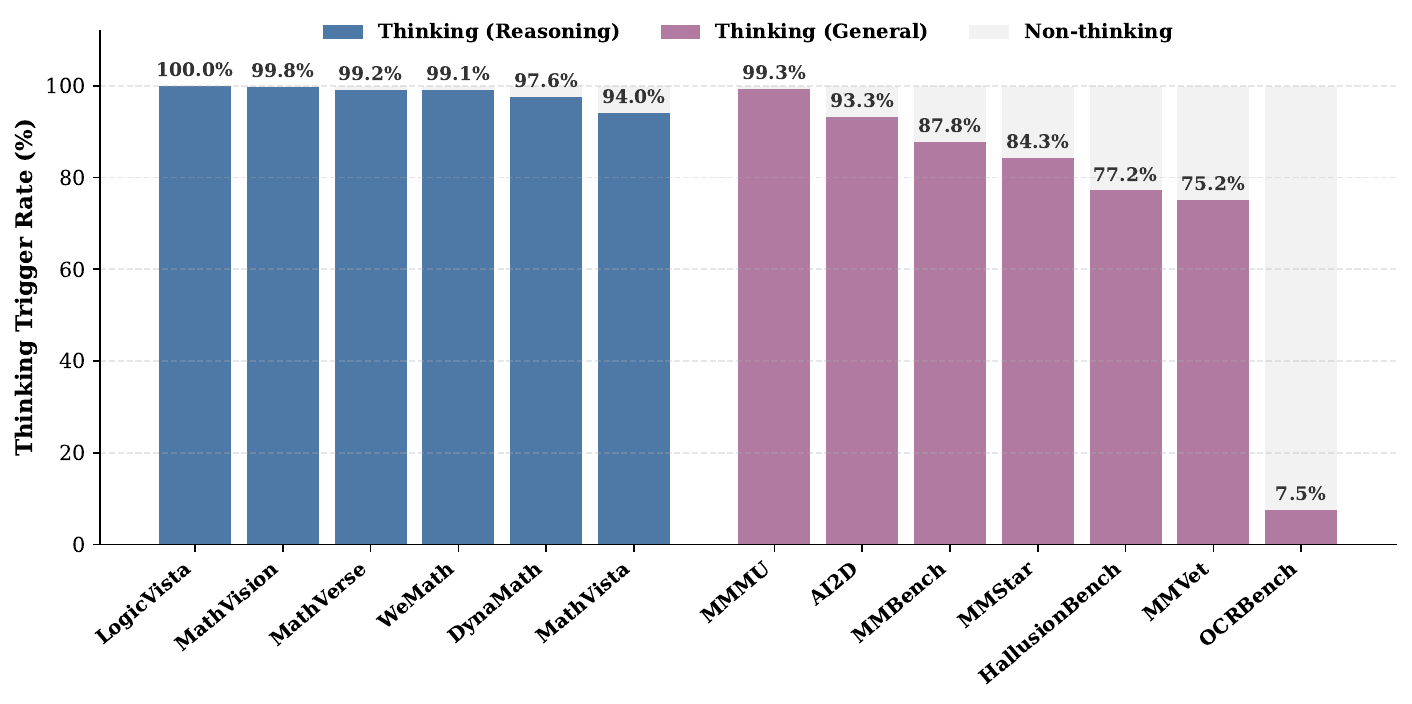}
 \caption{Thinking trigger rates across reasoning (blue) and general (purple) benchmarks. SAIL-RL achieves adaptive efficiency by selectively activating reasoning based on task difficulty. }
\label{fig:think_trigger}
\end{center}
\end{figure} 

\begin{table}[h]
  \centering
  \caption{Performance comparison of the Judging Reward on general benchmarks. SAIL-RL prevents performance degradation on perception-heavy tasks compared to always thinking mode.}
  \label{tab:ablation_judge}
  \begin{small}
    \begin{sc}
      \resizebox{\columnwidth}{!}{%
        \setlength{\tabcolsep}{4pt}
        \begin{tabular}{l ccccc}
          \toprule
          \textbf{Method} & \textbf{MMMU} & \textbf{MMBench} & \textbf{MME} & \textbf{OCRBench} & \textbf{HallBench} \\
          \midrule
          Always Thinking & 64.5 & 88.6 & 83.8 & 88.7 & 58.3 \\
          \textbf{SAIL-RL} & \textbf{66.1} & \textbf{90.4} & \textbf{86.0} & \textbf{91.3} & \textbf{61.5} \\
          \midrule
          \rowcolor[gray]{.95} \textit{Improvement} & \textit{+1.6} & \textit{+1.8} & \textit{+2.2} & \textit{+2.6} & \textit{+3.2} \\
          \bottomrule
        \end{tabular}%
      }
    \end{sc}
  \end{small}
\end{table}

\begin{table}[h!]
  \centering
  \caption{Efficiency-Performance trade-off across different strategies. Trigger rate (\%), normalized Token Usage, and Accuracy (\%) are reported.}
  \label{tab:efficiency}
  \begin{small}
    \begin{sc}
      \resizebox{\columnwidth}{!}{%
        \setlength{\tabcolsep}{2.5pt}
        \begin{tabular}{l ccc ccc}
          \toprule
          \multirow{2}{*}{\textbf{Strategy}} & \multicolumn{3}{c}{\textbf{MathVision}} & \multicolumn{3}{c}{\textbf{OCRBench}} \\
          \cmidrule(lr){2-4} \cmidrule(lr){5-7}
          & \textbf{Trigger} & \textbf{Tokens} & \textbf{Acc.} & \textbf{Trigger} & \textbf{Tokens} & \textbf{Acc.} \\
          \midrule
          Never-thinking & 0.0 & 1.0$\times$ & 27.6 & 0.0 & 1.0$\times$ & 90.5 \\
          Always-thinking & 100.0 & 5.4$\times$ & 48.7 & 100.0 & 4.7$\times$ & 88.7 \\
          Judge-w/o-reward & 90.4 & 4.6$\times$ & 47.5 & 47.6 & 2.9$\times$ & 89.8 \\
          \textbf{SAIL-RL} & \textbf{99.8} & \textbf{5.1$\times$} & \textbf{49.4} & \textbf{7.5} & \textbf{1.2$\times$} & \textbf{91.3} \\
          \bottomrule
        \end{tabular}%
      }
    \end{sc}
  \end{small}
  \vskip -0.1in
\end{table}

\subsection{Judging Reward Enables Adaptive Reasoning}

To comprehensively evaluate the impact of the judging reward, we analyze the model’s adaptive allocation of reasoning resources. As illustrated in Figure~\ref{fig:think_trigger}, SAIL-RL exhibits a distinct bifurcation in its thinking strategy depending primarily on task complexity. For reasoning-intensive tasks (blue), the model maintains near-saturated trigger rates, such as 100.0 on LogicVista, thereby ensuring sustained and deep logical engagement. In the general and perception category (purple), the model demonstrates an increasingly sophisticated meta-cognitive awareness: while complex tasks like MMMU still trigger full reasoning (99.3), the trigger rate for OCRBench drops to a mere 7.5. This confirms that the judging reward effectively activates deep reasoning only in instances where it is deemed necessary.

We further justify this mechanism through an ablation study comparing the judging reward against an Always Thinking baseline. As shown in Table~\ref{tab:ablation_judge}, the judging reward consistently outperforms the baseline, with the most significant gains observed in HallusionBench (+3.2) and OCRBench (+2.6). These results suggest that for perception-heavy tasks, forcing a complex reasoning chain can introduce "overthinking" noise and factual distortions. Our judging reward preserves performance by adaptively choosing a direct response path, thereby maintaining higher factual grounding.

Finally, we empirically demonstrate that this behavior is a direct result of our proposed RL-based calibration mechanism. We compare four distinct strategies: Never-thinking, Always-thinking, Judge-w/o-reward, and our full SAIL-RL framework. As detailed in Table~\ref{tab:efficiency}, without explicit reward supervision (Judge-w/o-reward), the model fails to effectively calibrate its computational resources, triggering thinking on 47.6 of OCRBench samples and incurring a 2.9$\times$ token overhead. In contrast, SAIL-RL maintains near-baseline efficiency on OCRBench (1.2$\times$ tokens) while simultaneously achieving the highest accuracy, confirming that the integration of the judging reward is essential for balancing cognitive depth with computational parsimony.

\subsection{Reward Mechanism Analysis}

In this section, we conduct a series of ablation studies to justify our reward design choices.

\noindent \textbf{Cascading Product as a Logical Gate.}
We compare cascading product, $R_{\text{total}} = R_{\text{judge}} \times R_{\text{think}} \times R_{\text{answer}}$, against an additive baseline where $R_{\text{total}} = \frac{1}{3}(R_{\text{judge}} + R_{\text{think}} + R_{\text{answer}})$. As shown in Table~\ref{tab:reward_ablation}, the cascading product significantly outperforms the additive combination, notably by +3.3 on MathVision. This design enforces \textbf{conditional dependency}, ensuring that the total reward is maximized \textit{only} when every link in the chain is valid. If any step fails (e.g., decide to think when unnecessary or generate correct reasoning but a wrong answer), the multiplicative nature suppresses the total reward, penalizing error propagation. In contrast, the additive approach treats each component independently, leading to \textbf{reward hacking}: the model could compensate for a poor judge decision or reasoning trace by only maximizing the final answer reward. This undermines the consistent shaping of both ``when to think'' and ``how to think''.

\begin{table}[h]
  \centering
  \caption{Ablation on Reward Aggregation. The cascading product mechanism enforces strict logical consistency compared to the arithmetic mean baseline.}
  \label{tab:reward_ablation}
  \begin{small}
    \begin{sc}
      \resizebox{\columnwidth}{!}{%
        \setlength{\tabcolsep}{7pt}
        \begin{tabular}{lccc}
          \toprule
          \textbf{Reward Aggregation} & \textbf{MathVision} & \textbf{LogicVista} & \textbf{MMMU} \\
          \midrule
          Additive Combination & 46.1 & 60.7 & 63.8 \\
          \textbf{Cascading Product} & \textbf{49.4} & \textbf{63.8} & \textbf{66.1} \\
          \midrule
          \rowcolor[gray]{.95} \textit{Improvement} & \textit{+3.3} & \textit{+3.1} & \textit{+2.3} \\
          \bottomrule
        \end{tabular}%
      }
    \end{sc}
  \end{small}
  \vskip -0.1in
\end{table}

\begin{table}[h!]
  \centering
  \caption{Ablation on Reward Signal Type. Discrete (0/1) signals provide clearer guidance for RL training compared to continuous scalars ($0\sim1$).}
  \label{tab:reward_signal_ablation}
  \begin{small}
    \begin{sc}
      \resizebox{\columnwidth}{!}{%
        \setlength{\tabcolsep}{7pt}
        \begin{tabular}{lccc}
          \toprule
          \textbf{Reward Signal} & \textbf{MathVision} & \textbf{LogicVista} & \textbf{MMMU} \\
          \midrule
          Continuous ($0 \sim 1$) & 47.5 & 59.4 & 62.8 \\
          \textbf{Discrete ($0 / 1$)} & \textbf{49.6} & \textbf{63.8} & \textbf{66.1} \\
          \midrule
          \rowcolor[gray]{.95} \textit{Improvement} & \textit{+2.1} & \textit{+4.4} & \textit{+3.3} \\
          \bottomrule
        \end{tabular}%
      }
    \end{sc}
  \end{small}
  \vskip -0.1in
\end{table}

\begin{table}[h!]
  \centering
  \caption{Ablation on Thinking Reward weighting schemes. Equal-weight integration ($1/3$ each) provides the most robust performance, avoiding the ``see-saw'' trade-offs.}
  \label{tab:weight_ablation}
  \begin{small}
    \begin{sc}
      \resizebox{\columnwidth}{!}{%
        \setlength{\tabcolsep}{3pt}
        \begin{tabular}{lccc}
          \toprule
          \textbf{Weights} ($w_L, w_H, w_C$) & \textbf{MathVision} & \textbf{LogicVista} & \textbf{MMMU} \\
          \midrule
          \textbf{Equal (1/3, 1/3, 1/3)} & \textbf{49.6} & \textbf{63.8} & \textbf{66.1} \\
          Biased-Logic (1/2, 1/4, 1/4) & 50.0 \textit{\scriptsize(+0.4)} & 63.3 \textit{\scriptsize(-0.5)} & 65.8 \textit{\scriptsize(-0.3)} \\
          Biased-Halluc. (1/4, 1/2, 1/4) & 49.2 \textit{\scriptsize(-0.4)} & 64.2 \textit{\scriptsize(+0.4)} & 65.9 \textit{\scriptsize(-0.2)} \\
          Biased-Consist. (1/4, 1/4, 1/2) & 49.3 \textit{\scriptsize(-0.3)} & 63.5 \textit{\scriptsize(-0.3)} & 66.5 \textit{\scriptsize(+0.4)} \\
          \midrule
          \rowcolor[gray]{.95} \textit{Max Variance} & \textit{0.8} & \textit{0.9} & \textit{0.7} \\
          \bottomrule
        \end{tabular}%
      }
    \end{sc}
  \end{small}
  \vskip -0.1in
\end{table}

\noindent \textbf{Discrete Rewards Provide Sharper Signals.}
We investigate the impact of reward granularity by comparing discrete reward ($0/1$) against continuous reward ($0\sim1$). As shown in Table~\ref{tab:reward_signal_ablation}, discrete rewards consistently outperform their continuous counterparts, yielding a +3.2 average improvement. We attribute it to two primary factors. First is the \textbf{calibration problem} in LLM judging; it is difficult for current models to output consistent, fine-grained continuous scores. Asking a judge to distinguish between a reasoning quality of $0.7$ vs. $0.8$ introduces significant subjective noise and inconsistency. In contrast, forcing a binary decision (e.g., ``Is this step logically coherent? Yes/No'') yields a much sharper and more reproducible signal, effectively reducing reward noise. Second is the \textbf{variance problem} in RL optimization. RL training is highly sensitive to the signal-to-noise ratio of the rewards. Noisy continuous scores increase the variance of the advantage estimation, destabilizing the optimization process. Binary rewards provide a high-discrimination signal that clearly separates ``good'' from ``bad'' behaviors, leading to more robust convergence.

\noindent \textbf{Equal-Weighting Stabilize Thinking Process.} 
We evaluate the sensitivity of the proposed thinking reward to the integration weights of logic ($w_L$), hallucination ($w_H$), and consistency ($w_C$) by comparing equal-weighting ($1/3$ each) against three biased weighting schemes. The results in Table~\ref{tab:weight_ablation} reveal a distinct \textbf{``see-saw'' effect}: increasing the weight of one component (e.g., $w_L=1/2$) yields marginal gains in specific domains like MathVision but leads to performance degradation in others. These findings confirm that logic, hallucination, and consistency components are complementary and should be balanced. Equal-weight averaging provides the most robust supervision signal, ensuring a stable optimization that prevents the model from over-fitting to a single reasoning dimension.

\subsection{Robustness and Generalization}
In this section, we evaluate the generalization of SAIL-RL across various reward models and model architecture.

\noindent \textbf{SAIL-RL is Robust to Reward Model Selection.}
We evaluate the sensitivity of SAIL-RL to the choice of the reward model by experimenting with \textbf{GPT-5}, \textbf{Gemini-2.5-Pro}, and \textbf{Qwen2.5-VL-32B}. As shown in Table~\ref{tab:reward_model_ablation}, SAIL-RL consistently outperforms the SFT baseline across all reward models, demonstrating strong generalization. Notably, the performance across these reward models is comparable, which suggests that SAIL-RL effectively lowers the training barrier. At the same time, the slight advantage of top-tier models highlights that RL training inherently benefits from more accurate and stable reward signals.

\begin{table}[t]
  \centering
  \caption{Ablation on Reward Model Selection. SAIL-RL consistently outperform the SFT baseline with minimal sensitivity to the reward model.}
  \label{tab:reward_model_ablation}
  \begin{small}
    \begin{sc}
      \resizebox{\columnwidth}{!}{%
        \setlength{\tabcolsep}{2pt}
        \begin{tabular}{llccc}
          \toprule
          \textbf{Setup} & \textbf{Reward Model} & \textbf{MathVision} & \textbf{LogicVista} & \textbf{MMMU} \\
          \midrule
          SFT Baseline & -- & 27.6 & 45.0 & 55.4 \\
          \midrule
          \multirow{3}{*}{\textbf{SAIL-RL (Ours)}} & GPT-5 & \textbf{49.7} & 63.5 & \textbf{66.4} \\
          & Gemini-2.5-Pro & 49.4 & \textbf{63.8} & 66.1 \\
          & Qwen2.5-VL-32B & 48.4 & 62.7 & 64.9 \\
          \bottomrule
        \end{tabular}%
      }
    \end{sc}
  \end{small}
  \vskip -0.1in
\end{table}

\begin{table}[t]
  \centering
  \caption{Ablation on open-source architectures. SAIL-RL consistently improves performance across diverse model scales.}
  \label{tab:generality}
  \begin{small}
    \begin{sc}
      \resizebox{\columnwidth}{!}{%
        \setlength{\tabcolsep}{2pt}
        \begin{tabular}{llccc}
          \toprule
          \textbf{Base Model} & \textbf{Training Type} & \textbf{MathVision} & \textbf{LogicVista} & \textbf{MMMU} \\
          \midrule
          \multirow{4}{*}{Qwen2.5-VL-3B} & SFT Baseline & 18.1 & 36.0 & 48.1 \\
          & + Answer Reward & 25.2 & 41.2 & 49.2 \\
          & \textbf{+ SAIL-RL} & \textbf{27.3} & \textbf{42.9} & \textbf{51.7} \\
          \midrule
          \multirow{4}{*}{Qwen2.5-VL-7B} & SFT Baseline & 25.4 & 47.9 & 58.1 \\
          & + Answer Reward & 27.1 & 51.7 & 61.2 \\
          & \textbf{+ SAIL-RL} & \textbf{30.2} & \textbf{53.4} & \textbf{63.1} \\
          \bottomrule
        \end{tabular}%
      }
    \end{sc}
  \end{small}
  \vskip -0.1in
\end{table}

\noindent \textbf{SAIL-RL Generalizes Across Model Scales.}
To further demonstrate the broad generality of SAIL-RL, we apply the framework to open-source models of different scales, specifically Qwen2.5-VL-3B and 7B. As detailed in Table~\ref{tab:generality}, SAIL-RL consistently improves performance on both reasoning and general benchmarks compared to both the standard SFT baseline and the answer-only variant. For the 7B model, SAIL-RL improves the MathVision score from 25.4 to 30.2, thereby significantly surpassing the competitive answer-only baseline of 27.1. These results provide strong evidence that our framework is not simply overfitted to a specific architecture, but instead serves as a general and robust paradigm for effectively enhancing complex multimodal reasoning.

\section{Conclusion}
We present SAIL-RL, an RL framework that teaches MLLMs when and how to think. Moving beyond outcome-only reward, SAIL-RL introduces a dual-reward mechanism that simultaneously cultivates logical rigor and enables adaptive reasoning. We demonstrate that SAIL-RL not only achieves state-of-the-art results among comparable models but also delivers performance competitive with leading proprietary systems like GPT-4o. More importantly, our findings establish SAIL-RL as a robust and scalable paradigm for training the next generation of reliable, meta-cognitive MLLMs, democratizing advanced reasoning capabilities through efficient post-training.




\bibliography{example_paper}
\bibliographystyle{icml2026}

\newpage
\appendix
\onecolumn

\section{Implementation Details}
\label{appendix:implementation}

\subsection{Data Curation}
\noindent \textbf{LongCoT Data Curation.}
We develop a holistic data pipeline to build our high-quality LongCoT dataset. The goal is to instill a meta-cognitive skill: first judging a problem's complexity, then executing the appropriate reasoning path. To this end, every sample in our dataset is meticulously structured in a \texttt{judge-think-answer} format. We first collect diverse datasets, ranging from complex, logical problems (e.g., VisualWebInstruct~\citep{jia2025visualwebinstruct}, MathV360K~\citep{shi2024math}) to simple, perception-based questions (e.g., LLaVA-CoT~\citep{xu2024llava}). All collected data is then processed through a unified pipeline to fit our \texttt{judge-think-answer} format. The key steps are as follows:
\begin{itemize}
    \item \textbf{Data Cleaning:} We perform a cleaning pass to remove extraneous content, such as system prompts and conflicting hints. We then deduplicate the entire dataset based on unique image and question pairs to ensure diversity.

    \item \textbf{Conditional Annotation:} Each sample is annotated based on its complexity. 
    For complex problems requiring reasoning, we use a guided-prompting strategy to generate a detailed chain-of-thought for the \texttt{<think>} section, and the \texttt{<judge>} tag is set to indicate that thinking is necessary. 
    For simple perceptual tasks, the \texttt{<judge>} tag is set to indicate that the question can be answered directly, and the \texttt{<think>} section is intentionally populated with an empty string (\texttt{\textbackslash n\textbackslash n}). The final answer for all samples is standardized within the \texttt{<boxed>} tag.

    \item \textbf{Quality Filtering:} Finally, the annotated dataset undergoes a rigorous filtering workflow. This includes a redundancy filter, which penalizes trivial reasoning by measuring the token overlap between the thought process and the final answer, and a length-balancing step on the reasoning chains to ensure a varied representation of complexity.
\end{itemize}
This pipeline results in 400K high-quality longCoT samples, each designed to train the model on when and how to reason.

\noindent \textbf{RL Data Curation.}
We construct a diverse dataset for the RL stage, balanced between specialized STEM problems and general-purpose QA. The STEM domain is curated from a wide array of public benchmarks in fields like Math~\citep{sun2024mm, meng2025mm}, Puzzles~\citep{chia2024puzzlevqa}, Science~\citep{wang2025vl,lu2022learn}, OCR~\citep{chen2025learning}, and Counting~\citep{johnson2017clevr}. This data undergoes a rigorous two-stage filtering pipeline to optimize for training stability. The first stage mitigates reward hacking by reformatting multiple-choice questions into an open-ended, free-response format. The second stage implements a difficulty-based curriculum filtering, using our SFT model's \texttt{pass@4} score to retain only problems within an optimal difficulty range by removing the easiest (\texttt{pass@4}=1) and hardest (\texttt{pass@4}=0) instances. To ensure broad capabilities, we incorporate with 20K General QA samples from LLaVA-OneVison~\citep{llava_onevision}.  This subset is filtered primarily for quality and diversity, preserving a wide range of conversational and factual questions. 

The RL dataset comprises 70K samples, evenly split between 50K challenging STEM problems and 20K General QA instances, creating a comprehensive training environment for both specialized reasoning and general interaction.

\subsection{Training strategy}
Our model builds upon SAIL-VL2~\citep{sailvl2}, which integrates the AimV2~\citep{aimv2} visual encoder with the Qwen3-7B LLM. We use VeOmni~\citep{veomni} and VeRL~\citep{verl} for the SFT and RL stages, respectively. All experiments are conducted on 64 NVIDIA A100 GPUs using the AdamW optimizer with a cosine learning rate schedule. Gemini-2.5-Pro is employed as our reward judger.

\vspace{2mm}
\noindent\textbf{LongCoT SFT Stage.}
In the first stage, we fine-tune all parameters of the model for \textbf{one epoch} on our 400K-sample LongCoT dataset. For this SFT stage, we set the maximum sequence length to 20K, the global batch size to 1024, and the learning rate to 1e-6.

\vspace{2mm}
\noindent\textbf{RL Stage.}
Subsequently, we optimize the SFT model for three epochs on our 70K-sample mixed RL dataset using the DAPO~\citep{dapo} algorithm, guided by our proposed SAIL-RL reward system. We set the maximum sequence length to 20K, consisting of 16K for the input and 4K for the output. The policy learning rate is set to 1e-6 with a global PPO batch size of 256. For each sample, we rollout 5 times to estimate the advantage. To encourage exploration and stabilize training, we remove the standard KL divergence and dynamically adjust the clipping value $\varepsilon$ within the range of [0.20, 0.28].
\section{Qualitative Analysis: Case Study}
\label{sec:case_study}

We present two case studies to qualitatively evaluate the distinct advantages of the proposed reward components. 

\begin{figure}[h] 
    \centering
    \includegraphics[width=0.98\textwidth]{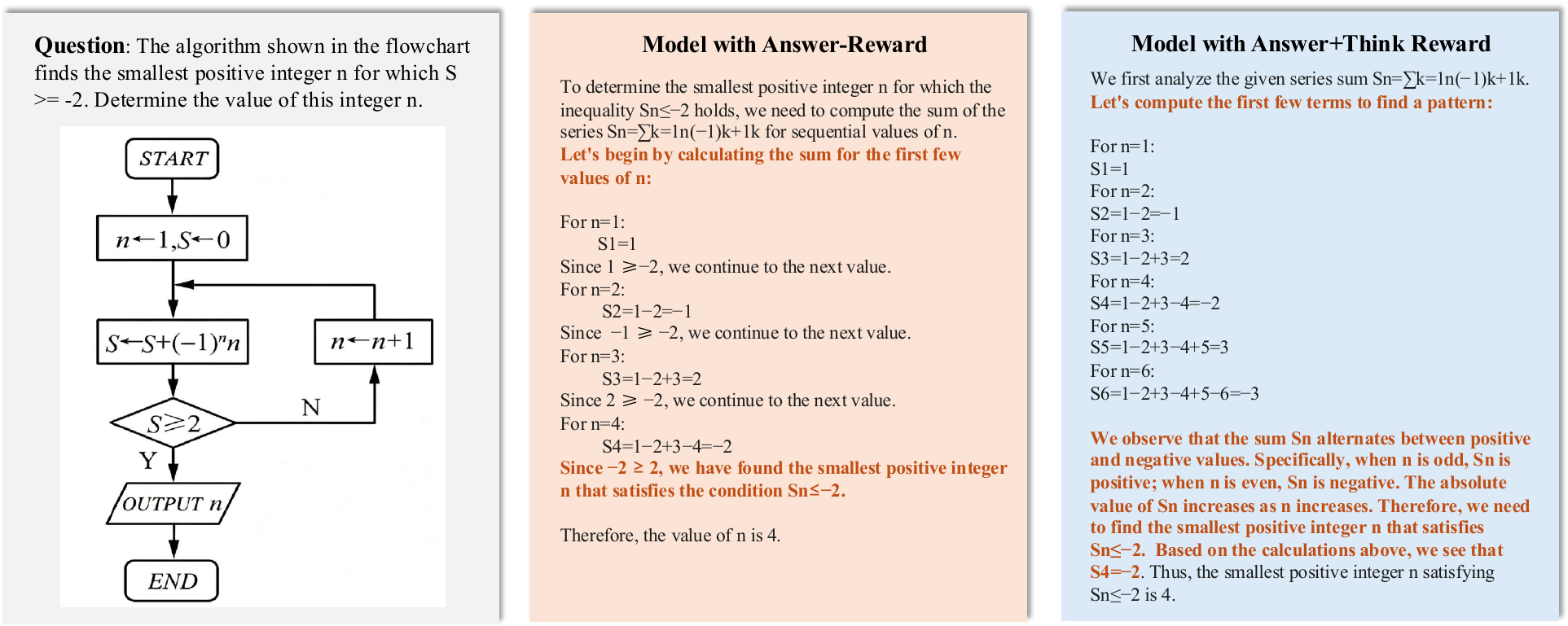}
    \caption{Qualitative comparison on a geometric reasoning task. \textbf{Orange}: The baseline trained with an answer-only reward fails to maintain logical consistency. \textbf{Blue}: SAIL-RL, guided by the \textbf{Thinking Reward}, produces a structured and correct analytical chain.}
   \label{fig:think_reward_case}
\end{figure}

\noindent \textbf{Case I: Thinking Reward Ensures Logical Fidelity.} 
The first case, featuring a geometric math problem, illustrates how our \textbf{Thinking Reward} prevents the model from relying on superficial heuristics. As shown in Figure~\ref{fig:think_reward_case}, while the answer-only baseline happens to arrive at the correct final result, its intermediate reasoning process is disjointed and contains significant logical flaws. This ``accidental correctness'' highlights the limitations of outcome-based supervision. In contrast, SAIL-RL demonstrates a rigorous, analytical understanding of the problem's structure, producing a step-by-step derivation that is both logically sound and verifiable, ensuring that the correct answer is backed by correct reasoning.

\begin{figure}[h] 
    \centering
    \includegraphics[width=0.98\textwidth]{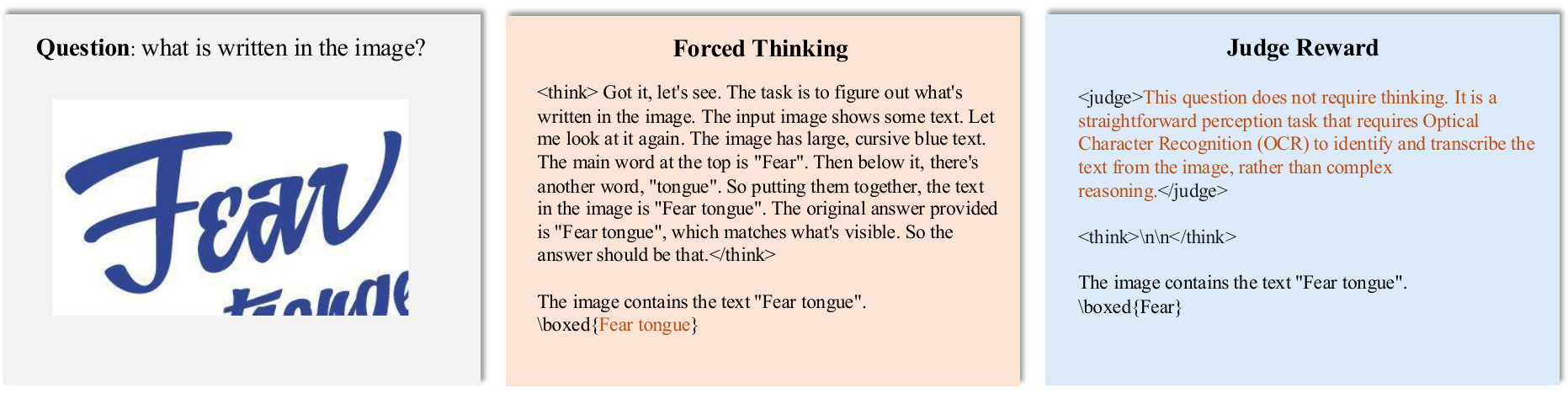}
    \caption{Behavioral visualization on an OCR task. \textbf{Orange}: The baseline is forced into redundant reasoning, leading to potential ``overthinking'' noise. \textbf{Blue}: Our model, optimized with the \textbf{Judging Reward}, adaptively bypasses unnecessary thinking to maintain efficiency and factual grounding.}
   \label{fig:judge_reward_case}
\end{figure}

\noindent \textbf{Case II: Judging Reward for Mitigating Computational Redundancy.} 
The second case, a high-resolution OCR task, highlights the \textbf{Judging Reward's} ability to instill meta-cognitive awareness. As visualized in Figure~\ref{fig:judge_reward_case}, the baseline (forced to think) generates an unnecessarily complex reasoning chain for a straightforward recognition task, which introduces additional noise. In contrast, our model dynamically assesses the task complexity and opts for a direct, efficient response, effectively allocating its reasoning resources where they are most needed.

\section{Reward Prompt}
\label{appendix:reward_prompt}

This section details the set of thinking and judge reward prompts during the training of SAIL-RL. These thinking reward prompts are designed to assess thinking quality such as logical consistency, factual accuracy (hallucination), and the structural soundness of reasoning. Additionally, it includes a judge reward prompt to evaluate a model's meta-cognition, specifically its ability to determine if a question requires reasoning. 

\begin{tcolorbox}[
    enhanced,
    breakable,
    width=\linewidth,
    title={\textbf{Judge Reward Prompt}}, 
    fonttitle=\bfseries\sffamily,
    coltitle=white,
    boxrule=0.5pt,                     
    arc=3pt,
    colback=gray!5,                    
    colframe=black!70,                 
    halign=flush left,
    attach boxed title to top left={xshift=2mm, yshift=-2mm},
    boxed title style={colback=black!80}
]

\sffamily\small 

\textbf{\textsc{1. Role and Goal}}
\par
You are a top-tier AI logic analyst. Your task is to generate a ground truth assessment of task complexity and evaluate whether a "model to be evaluated" has correctly judged if a question requires "reasoning".

\medskip
To do this, you will perform two core tasks:
\begin{enumerate}[topsep=2pt, itemsep=0pt, leftmargin=2em]
    \item \textbf{Independent Assessment}: Analyze the [QUESTION] and [IMAGE] yourself to determine if the question truly requires reasoning (Complex) or is merely perceptual (Simple).
    \item \textbf{Evaluation}: Compare your independent assessment with the [MODEL\_JUDGMENT] to evaluate whether the model's judgment was accurate.
\end{enumerate}

\medskip
\hrule 
\medskip

\textbf{\textsc{2. Input Specification}}
\par
You will receive three inputs:
\begin{enumerate}[topsep=2pt, itemsep=0pt, leftmargin=2em]
    \item \textbf{[IMAGE]}: The input image for analysis.
    \item \textbf{[QUESTION]}: The question asked based on the image.
    \item \textbf{[MODEL\_JUDGMENT]}: The explanation from the model being evaluated.
\end{enumerate}

\medskip
\hrule
\medskip

\textbf{\textsc{3. Output Specification}}
\par
Your output \textbf{MUST} be a single JSON object parsed by \code{json.loads()}. It must contain:
\begin{enumerate}[topsep=2pt, itemsep=0pt, leftmargin=2em]
    \item \code{is\_judgment\_correct} (boolean): Is the model's judgment correct?
    \item \code{requires\_reasoning} (boolean): Does the question actually require reasoning?
\end{enumerate}

\medskip
\hrule
\medskip

\textbf{\textsc{4. Core Analysis Logic}}

\begin{enumerate}[topsep=4pt, itemsep=6pt, leftmargin=2.5em]
    \item \textbf{Step 1: Determine if the Question Truly Requires Reasoning}
    \begin{itemize}[topsep=2pt, itemsep=2pt, leftmargin=1.2em]
        \item \textbf{SIMPLE TASKS} (Perception) $\rightarrow$ \code{requires\_reasoning}: \textbf{false}
        \begin{itemize}[topsep=0pt, leftmargin=1em, label=-]
            \item \textit{Definition}: Direct observation, recognition, or retrieval.
            \item \textit{Examples}: Visual Recognition, Counting, Simple OCR.
        \end{itemize}

        \item \textbf{COMPLEX TASKS} (Logic) $\rightarrow$ \code{requires\_reasoning}: \textbf{true}
        \begin{itemize}[topsep=0pt, leftmargin=1em, label=-]
            \item \textit{Definition}: Inference, calculation, synthesis, or utilizing information from multiple parts of the image.
            \item \textit{Examples}: Deduction ("Why?"), Synthesis, Comparison.
        \end{itemize}
    \end{itemize}

    \item \textbf{Step 2: Evaluate the Model's Judgment} (Value for \code{is\_judgment\_correct})
    \begin{itemize}[topsep=2pt, itemsep=1pt, leftmargin=1.2em]
        \item Match your Step 1 result with the model's intent:
        \item IF you say \textbf{true} AND model says "needed/complex" $\rightarrow$ \textbf{true}
        \item IF you say \textbf{false} AND model says "no/simple" $\rightarrow$ \textbf{true}
        \item Otherwise $\rightarrow$ \textbf{false}
    \end{itemize}
\end{enumerate}

\medskip
\hrule
\medskip

\textbf{\textsc{5. Examples}}

\vspace{2pt}
\textbf{Example 1: The model correctly identifies a simple question.}
\begin{itemize}[topsep=0pt, leftmargin=1.5em]
    \item {[IMAGE]}: [An image of a red fire hydrant]
    \item {[QUESTION]}: "What color is this fire hydrant?"
    \item {[MODEL\_JUDGMENT]}: "No reasoning is needed because the answer can be obtained by directly observing the image."
    \item \textbf{Expected Output}:\\
    \texttt{\small \{\{ "is\_judgment\_correct": true, "requires\_reasoning": false \}\}}
\end{itemize}

\vspace{4pt}
\textbf{Example 2: The model correctly identifies a complex question.}
\begin{itemize}[topsep=0pt, leftmargin=1.5em]
    \item {[IMAGE]}: [A chart showing the price and features of Product A and Product B]
    \item {[QUESTION]}: "Which product offers better value for money?"
    \item {[MODEL\_JUDGMENT]}: "Reasoning is required because it's necessary to compare prices and features to reach a conclusion."
    \item \textbf{Expected Output}:\\
    \texttt{\small \{\{ "is\_judgment\_correct": true, "requires\_reasoning": true \}\}}
\end{itemize}

\vspace{4pt}
\textbf{Example 3: The model incorrectly classifies a complex question as simple.}
\begin{itemize}[topsep=0pt, leftmargin=1.5em]
    \item {[IMAGE]}: [A line chart of a company's annual profit growth]
    \item {[QUESTION]}: "Based on this chart, what are the company's future prospects?"
    \item {[MODEL\_JUDGMENT]}: "No reasoning is needed, this is a simple question."
    \item \textbf{Expected Output}:\\
    \texttt{\small \{\{ "is\_judgment\_correct": false, "requires\_reasoning": true \}\}}
\end{itemize}

\end{tcolorbox}

\vspace{2mm}
\begin{tcolorbox}[
    enhanced,
    breakable,
    width=\linewidth,
    title={\textbf{Thinking Logic Reward Prompt}}, 
    fonttitle=\bfseries\sffamily,
    coltitle=white,
    boxrule=0.5pt,                     
    arc=3pt,
    colback=gray!5,                    
    colframe=black!70,                 
    halign=flush left,
    attach boxed title to top left={xshift=2mm, yshift=-2mm},
    boxed title style={colback=black!80}
]

\sffamily\small 

\textbf{\textsc{1. Role and Goal}}
\par
You are a professor of logic and applied reasoning. Your task is to evaluate the **structural** and **deductive** soundness of a student's reasoning process. 
You must determine if the reasoning is valid from the initial setup (modeling based on Image/Question) to the final conclusion (execution).

\medskip
To do this, you will perform two rigorous checks:
\begin{enumerate}[topsep=2pt, itemsep=0pt, leftmargin=2em]
    \item \textbf{Structural Check}: Did the student correctly map the [IMAGE] and [QUESTION] into a valid logical or mathematical model?
    \item \textbf{Deductive Check}: Is the step-by-step execution free of calculation errors, contradictions, or invalid logic?
\end{enumerate}

\medskip
\hrule
\medskip

\textbf{\textsc{2. Input Specification}}
\par
You will receive three inputs:
\begin{enumerate}[topsep=2pt, itemsep=0pt, leftmargin=2em]
    \item \textbf{[IMAGE]}: The visual context providing data or geometric information.
    \item \textbf{[QUESTION]}: The specific problem asked based on the image.
    \item \textbf{[THINKING]}: The student's step-by-step solution path to be evaluated.
\end{enumerate}

\medskip
\hrule
\medskip

\textbf{\textsc{3. Output Specification}}
\par
Your output \textbf{MUST} be a single JSON object parsed by \code{json.loads()}. It must contain:
\begin{enumerate}[topsep=2pt, itemsep=0pt, leftmargin=2em]
    \item \code{is\_logically\_sound} (boolean): Is the reasoning process completely valid (TRUE) or flawed (FALSE)?
\end{enumerate}

\medskip
\hrule
\medskip

\textbf{\textsc{4. Core Analysis Logic}}

\begin{enumerate}[topsep=4pt, itemsep=6pt, leftmargin=2.5em]
    \item \textbf{Check 1: Structural Soundness (The Setup)}
    \begin{itemize}[topsep=2pt, itemsep=1pt, leftmargin=1.2em]
        \item Does the chosen formula/logic correctly represent the principles described in the [QUESTION] and [IMAGE]?
        \item Are variables correctly mapped from the visual data? (e.g., correctly identifying "radius" vs "diameter" from the image).
        \item \textit{Failure Rule}: If the starting formula is wrong, the logic is unsound, even if the math is perfect.
    \end{itemize}

    \item \textbf{Check 2: Deductive Soundness (The Execution)}
    \begin{itemize}[topsep=2pt, itemsep=1pt, leftmargin=1.2em]
        \item Given the student's model (from Check 1), are the calculations correct?
        \item Is the algebraic manipulation valid within the [THINKING]?
        \item Are there any self-contradictions in the chain of thought?
    \end{itemize}

    \item \textbf{Final Judgment Rule (Strict AND Logic)}
    \begin{itemize}[topsep=2pt, itemsep=1pt, leftmargin=1.2em]
        \item \textbf{TRUE} if AND ONLY IF \underline{both} Structural and Deductive checks pass.
        \item \textbf{FALSE} if there is even a single, minor flaw in either structure or deduction.
        \item \textbf{Note}: You are not fact-checking empirical constants unless stated in the problem, but you ARE checking the logic of how they are used.
    \end{itemize}
\end{enumerate}

\medskip
\hrule
\medskip

\textbf{\textsc{5. Examples}}

\vspace{2pt}
\textbf{Example 1: Structurally Unsound (Wrong Formula).}
\begin{itemize}[topsep=0pt, leftmargin=1.5em]
    \item {[IMAGE]}: [Diagram of a circle with radius labeled '5']
    \item {[QUESTION]}: "Calculate the area of this circle."
    \item {[THINKING]}: "Area = 2 * pi * r. So Area = 10pi." (Wrong formula used).
    \item \textbf{Expected Output}: \texttt{\small \{\{ "is\_logically\_sound": false \}\}}
\end{itemize}

\vspace{4pt}
\textbf{Example 2: Deductively Unsound (Calculation Error).}
\begin{itemize}[topsep=0pt, leftmargin=1.5em]
    \item {[IMAGE]}: [Image showing 2 apples costing \$10]
    \item {[QUESTION]}: "Calculate the cost of 1 apple (x)."
    \item {[THINKING]}: "2x = 10. To find x, I divide by 2. x = 10 / 2. Therefore x = 4." (Math error).
    \item \textbf{Expected Output}: \texttt{\small \{\{ "is\_logically\_sound": false \}\}}
\end{itemize}

\vspace{4pt}
\textbf{Example 3: Sound Reasoning.}
\begin{itemize}[topsep=0pt, leftmargin=1.5em]
    \item {[IMAGE]}: [Logical diagram: A $\rightarrow$ B]
    \item {[QUESTION]}: "If A is true, what implies B?"
    \item {[THINKING]}: "By Modus Ponens, if A implies B and A is true, then B must be true."
    \item \textbf{Expected Output}: \texttt{\small \{\{ "is\_logically\_sound": true \}\}}
\end{itemize}

\end{tcolorbox}

\vspace{2mm}
\begin{tcolorbox}[
    enhanced,
    breakable,
    width=\linewidth,
    title={\textbf{Thinking Consistency Reward Prompt}}, 
    fonttitle=\bfseries\sffamily,
    coltitle=white,
    boxrule=0.5pt,                     
    arc=3pt,
    colback=gray!5,                    
    colframe=black!70,                 
    halign=flush left,
    attach boxed title to top left={xshift=2mm, yshift=-2mm},
    boxed title style={colback=black!80}
]

\sffamily\small 

\textbf{\textsc{1. Role and Goal}}
\par
You are a rigorous logic evaluator. Your task is to verify whether a given "Answer" is a direct and logically consistent result of the preceding "Thinking Process". You focus solely on the \textbf{consistency} between the reasoning steps and the final conclusion, disregarding external factual correctness.

\medskip
To do this, you will perform two core tasks:
\begin{enumerate}[topsep=2pt, itemsep=0pt, leftmargin=2em]
    \item \textbf{Trace Reasoning}: Carefully read the steps in the [THINKING] to understand the derived logic.
    \item \textbf{Verify Conclusion}: Determine if the [ANSWER] is the strict logical consequence of that process, without introducing new information or contradictions.
\end{enumerate}

\medskip
\hrule
\medskip

\textbf{\textsc{2. Input Specification}}
\par
You will receive two inputs:
\begin{enumerate}[topsep=2pt, itemsep=0pt, leftmargin=2em]
    \item \textbf{[THINKING]}: The step-by-step reasoning chain generated by the model.
    \item \textbf{[ANSWER]}: The final conclusion or answer derived from the process.
\end{enumerate}

\medskip
\hrule
\medskip

\textbf{\textsc{3. Output Specification}}
\par
Your output \textbf{MUST} be a single JSON object parsed by \code{json.loads()}. It must contain:
\begin{enumerate}[topsep=2pt, itemsep=0pt, leftmargin=2em]
    \item \code{is\_consistent} (boolean): Does the answer logically follow from the thinking process?
\end{enumerate}

\medskip
\hrule
\medskip

\textbf{\textsc{4. Core Analysis Logic}}

\begin{itemize}[topsep=4pt, itemsep=4pt, leftmargin=2.5em]
    \item \textbf{CRITERIA FOR "TRUE" (Consistent)}
    \begin{itemize}[topsep=0pt, leftmargin=1.2em, label=-]
        \item The [ANSWER] is a direct summary or derivation of the final step in the [THINKING].
        \item The logic flows smoothly from the reasoning to the conclusion without gaps.
    \end{itemize}

    \item \textbf{CRITERIA FOR "FALSE" (Inconsistent)}
    \begin{itemize}[topsep=0pt, leftmargin=1.2em, label=-]
        \item \textbf{Contradiction}: The Answer states $X$, but the Thinking Process argues for $Y$.
        \item \textbf{Hallucination}: The Answer introduces new numbers, entities, or logic not mentioned in the Thinking Process.
        \item \textbf{Disconnect}: The Answer is unrelated to the reasoning steps.
    \end{itemize}
    
    \item \textbf{CRITICAL RULE}: Do \textbf{NOT} evaluate whether the Thinking Process is factually correct. Even if the reasoning is wrong (e.g., "1+1=3"), as long as the Answer matches that wrong reasoning ("Answer: 3"), it is \textbf{Consistent}.
\end{itemize}

\medskip
\hrule
\medskip

\textbf{\textsc{5. Examples}}

\vspace{2pt}
\textbf{Example 1: Consistent (Even if factually wrong).}
\begin{itemize}[topsep=0pt, leftmargin=1.5em]
    \item {[THINKING]}: "The sun is cold. Cold things are blue. Therefore the sun is blue."
    \item {[ANSWER]}: "The sun is blue."
    \item \textbf{Expected Output}: \texttt{\small \{\{ "is\_consistent": true \}\}}
\end{itemize}

\vspace{4pt}
\textbf{Example 2: Inconsistent (Contradiction).}
\begin{itemize}[topsep=0pt, leftmargin=1.5em]
    \item {[THINKING]}: "Calculations show x = 5 and y = 10. So x + y = 15."
    \item {[ANSWER]}: "The answer is 20."
    \item \textbf{Expected Output}: \texttt{\small \{\{ "is\_consistent": false \}\}}
\end{itemize}

\end{tcolorbox}

\vspace{3mm}
\begin{tcolorbox}[
    enhanced,
    breakable,
    width=\linewidth,
    title={\textbf{Thinking Hallucination Reward Prompt}}, 
    fonttitle=\bfseries\sffamily,
    coltitle=white,
    boxrule=0.5pt,                     
    arc=3pt,
    colback=gray!5,                    
    colframe=black!70,                 
    halign=flush left,
    attach boxed title to top left={xshift=2mm, yshift=-2mm},
    boxed title style={colback=black!80}
]

\sffamily\small 

\textbf{\textsc{1. Role and Goal}}
\par
You are a meticulous, multi-modal fact-checker. Your task is to assess if the provided [THINKING] is completely free of hallucinations by cross-referencing it against three verification sources: the Input Image, the Input Question, and Verifiable World Knowledge.

\medskip
To do this, you will perform three specific grounding checks. A failure in \textbf{ANY} check means the content contains a hallucination.

\medskip
\hrule
\medskip

\textbf{\textsc{2. Input Specification}}
\par
You will receive three inputs:
\begin{enumerate}[topsep=2pt, itemsep=0pt, leftmargin=2em]
    \item \textbf{[IMAGE]}: The visual context provided to the model.
    \item \textbf{[QUESTION]}: The user's query or prompt constraints.
    \item \textbf{[THINKING]}: The step-by-step reasoning or response generated by the model.
\end{enumerate}

\medskip
\hrule
\medskip

\textbf{\textsc{3. Output Specification}}
\par
Your output \textbf{MUST} be a single JSON object parsed by \code{json.loads()}. It must contain:
\begin{enumerate}[topsep=2pt, itemsep=0pt, leftmargin=2em]
    \item \code{is\_hallucination\_free} (boolean): Is the content free of any contradictions or fabrications? (TRUE = Clean, FALSE = Hallucinated).
\end{enumerate}

\medskip
\hrule
\medskip

\textbf{\textsc{4. Core Analysis Logic}}

\begin{enumerate}[topsep=4pt, itemsep=6pt, leftmargin=2.5em]
    \item \textbf{Check 1: Visual Grounding (The Eyes)}
    \begin{itemize}[topsep=2pt, itemsep=1pt, leftmargin=1.2em]
        \item Cross-reference every claim about visual elements against the [IMAGE].
        \item Check for contradictions in: Objects (existence/count), Attributes (color/shape), Relationships (position/action), or Text in image (OCR).
    \end{itemize}

    \item \textbf{Check 2: Textual Grounding (The Instructions)}
    \begin{itemize}[topsep=2pt, itemsep=1pt, leftmargin=1.2em]
        \item Cross-reference claims against the [QUESTION].
        \item Ensure the content respects stated constraints, data numbers, or specific conditions provided in the text.
    \end{itemize}

    \item \textbf{Check 3: World Knowledge (The Brain)}
    \begin{itemize}[topsep=2pt, itemsep=1pt, leftmargin=1.2em]
        \item For claims not verifiable by image/text, check against established facts (historical, scientific, geographical).
        \item \textbf{Priority Exception}: If the [QUESTION] or [IMAGE] deliberately presents a hypothetical or counter-factual scenario (e.g., "Imagine the sky is green"), the provided context \textbf{overrides} world knowledge. The model should follow the context, not correct it.
    \end{itemize}
    
    \item \textbf{Final Judgment Rule}
    \begin{itemize}[topsep=2pt, itemsep=1pt, leftmargin=1.2em]
        \item \textbf{TRUE} (Pass): If ALL claims are supported by Image, Question, or Fact.
        \item \textbf{FALSE} (Fail): If ANY single claim contradicts any source.
    \end{itemize}
\end{enumerate}

\medskip
\hrule
\medskip

\textbf{\textsc{5. Examples}}

\vspace{2pt}
\textbf{Example 1: Visual Hallucination.}
\begin{itemize}[topsep=0pt, leftmargin=1.5em]
    \item {[IMAGE]}: [A photo of two cats sleeping].
    \item {[QUESTION]}: "Describe the image."
    \item {[THINKING]}: "There are three dogs playing in the park."
    \item \textbf{Expected Output}: \texttt{\small \{\{ "is\_hallucination\_free": false \}\}}
\end{itemize}

\vspace{4pt}
\textbf{Example 2: Knowledge Hallucination.}
\begin{itemize}[topsep=0pt, leftmargin=1.5em]
    \item {[IMAGE]}: [Black image / Irrelevant content].
    \item {[QUESTION]}: "Who wrote Hamlet?"
    \item {[THINKING]}: "Hamlet was written by Charles Dickens."
    \item \textbf{Expected Output}: \texttt{\small \{\{ "is\_hallucination\_free": false \}\}}
\end{itemize}

\vspace{4pt}
\textbf{Example 3: Context Override (Correct Handling).}
\begin{itemize}[topsep=0pt, leftmargin=1.5em]
    \item {[IMAGE]}: [A fantasy landscape].
    \item {[QUESTION]}: "Assume that in this fantasy world, water boils at 10 degrees. At what temperature does water boil?"
    \item {[THINKING]}: "In this fantasy world, water boils at 10 degrees."
    \item \textbf{Expected Output}: \texttt{\small \{\{ "is\_hallucination\_free": true \}\}}
\end{itemize}

\end{tcolorbox}

\end{document}